\documentclass{mva_style}
\usepackage{graphicx}

\usepackage{amssymb}
\usepackage{pifont}
\newcommand{\xmark}{\ding{55}}%
\usepackage{xcolor}
\definecolor{darkblue}{rgb}{0.0, 0.0, 0.55}
\usepackage{hyperref}

    \hypersetup{
         colorlinks = true,
         linkcolor =  darkblue,
         anchorcolor = blue,
         citecolor = darkblue,
         filecolor = blue,
         urlcolor = red
         }
\usepackage[T1]{fontenc}
\usepackage{color}
\usepackage{xcolor,colortbl}
\usepackage{xcolor}
\definecolor{aliceblue}{rgb}{0.94, 0.97, 1.0}
\definecolor{Gray}{gray}{0.9}
\definecolor{GrAy}{gray}{0.96}
\definecolor{beaublue}{rgb}{0.74, 0.83, 0.9}
\definecolor{LightCyan}{rgb}{0.88,1,1}
\definecolor{inchworm}{rgb}{0.7, 0.93, 0.36}
\definecolor{darksalmon}{rgb}{0.91, 0.59, 0.48}
\definecolor{babyblue}{rgb}{0.54, 0.81, 0.94}
\definecolor{jonquil}{rgb}{0.98, 0.85, 0.37}
\definecolor{green}{rgb}{0.58, 0.77, 0.45}
\definecolor{darksalmon}{rgb}{0.91, 0.59, 0.48}
\definecolor{bisque}{rgb}{1.0, 0.89, 0.77}
 \definecolor{inchworm}{rgb}{0.7, 0.93, 0.36}
\usepackage{comment}
\usepackage{xspace}
\usepackage{amsmath}
\usepackage{relsize}%
\usepackage{mathtools}
\usepackage{fontawesome}
\usepackage{adjustbox}
\usepackage{booktabs}
\usepackage{graphicx}
\usepackage{enumitem}
\usepackage{color}
\usepackage{latexsym}
\usepackage{graphicx}
\usepackage{amsmath}
\usepackage{mathtools}
\usepackage{breqn}
\usepackage{amsmath}
\usepackage{url}
\usepackage{enumerate}
\usepackage{amsmath}
\usepackage{amssymb}
\usepackage{comment}
\usepackage{xspace}
\usepackage{adjustbox}
\usepackage{booktabs}
\usepackage{graphicx}
\usepackage{enumitem}
\usepackage{color}
\usepackage[normalem]{ulem}
\usepackage{setspace}
\usepackage{relsize}

\usepackage{booktabs}
\usepackage{nicematrix}
\usepackage{xcolor}
\usepackage{color}
\usepackage{xcolor,colortbl}
\definecolor{aliceblue}{rgb}{0.94, 0.97, 1.0}
\definecolor{beaublue}{rgb}{0.74, 0.83, 0.9}
\definecolor{LightCyan}{rgb}{0.88,1,1}
\definecolor{inchworm}{rgb}{0.7, 0.93, 0.36}
\usepackage{comment}
\usepackage{xspace}
\usepackage{adjustbox}
\usepackage{booktabs}
\usepackage{graphicx}
\usepackage{enumitem}
\usepackage{color}
\usepackage{latexsym}
\usepackage{graphicx}
\usepackage{amsmath}
\usepackage{mathtools}
\usepackage{breqn}
\usepackage{amsmath}
\usepackage{url}
\usepackage{enumerate}
\usepackage{amsmath}
\usepackage{amssymb}
\usepackage{comment}
\usepackage{xspace}
\usepackage{adjustbox}
\usepackage{booktabs}
\usepackage{graphicx}
\usepackage{enumitem}
\usepackage{color}

\usepackage{setspace}
\usepackage{relsize}

\usepackage{booktabs}
\usepackage{nicematrix}
\usepackage{xcolor,colortbl}
\definecolor{aliceblue}{rgb}{0.94, 0.97, 1.0}
\definecolor{beaublue}{rgb}{0.74, 0.83, 0.9}
\definecolor{LightCyan}{rgb}{0.88,1,1}
\definecolor{inchworm}{rgb}{0.7, 0.93, 0.36}
\usepackage{comment}
\usepackage{xspace}
\usepackage{adjustbox}
\usepackage{booktabs}
\usepackage{graphicx}
\usepackage{enumitem}
\usepackage{color}
\usepackage{latexsym}
\usepackage{graphicx}
\usepackage{amsmath}
\usepackage{mathtools}
\usepackage{breqn}
\usepackage{amsmath}
\usepackage{url}
\usepackage{enumerate}
\usepackage{amsmath}
\usepackage{amssymb}
\usepackage{comment}
\usepackage{xspace}
\usepackage{adjustbox}
\usepackage{booktabs}
\usepackage{graphicx}
\usepackage{enumitem}
\usepackage{color}

\usepackage{setspace}
\usepackage{relsize}

\usepackage{booktabs}
\usepackage{nicematrix}
\usepackage{xcolor}

\usepackage{xcolor,colortbl}
\DeclareRobustCommand\onedot{\futurelet\@let@token\@onedot}
\def\@onedot{\ifx\@let@token.\else.\null\fi\xspace}

\def\eg{\emph{e.g}\onedot} 
\def\ie{\emph{i.e}\onedot}

\makeatother

\makeatletter
\DeclareRobustCommand\onedot{\futurelet\@let@token\@onedot}
\def\@onedot{\ifx\@let@token.\else.\null\fi\xspace}

\def\eg{\emph{e.g}\onedot} 
\def\ie{\emph{i.e}\onedot}

\makeatother

\newcolumntype{D}{ >{\vspace{-0.1000cm}\centering\arraybackslash} m{1cm} }
\usepackage{xcoffins}
\NewCoffin\tablecoffin

  \newcolumntype{C}[1]{>{\centering\let\newline\\\arraybackslash\hspace{0pt}}m{#1}}
\newcolumntype{L}[1]{>{\raggedright\let\newline\\\arraybackslash\hspace{0pt}}m{#1}}
\usepackage[utf8]{inputenc}

\usepackage{microtype}
\finalcopy 

\begin{document}
\title{ Word to Sentence Visual Semantic Similarity \\  for Caption Generation: Lessons Learned}

\author{Ahmed Sabir  \\
Universitat Polit\`ecnica de Catalunya, TALP Research Center, Barcelona, Spain }


\maketitle

\section*{\centering Abstract}
\textit{
This paper focuses on enhancing the captions generated by image captioning systems. We propose an approach for improving caption generation systems by choosing the most closely related output to the image rather than the most likely output produced by the model. Our model revises the language generation output beam search from a visual context perspective. We employ a visual semantic measure in a word and sentence level manner to match the proper caption to the related information in the image. This approach can be applied to any caption system as a post-processing method.}





\section{Introduction}

\vspace{-0.2cm}

\begin{figure}[t!]
\centering 

\includegraphics[width=0.8\columnwidth]{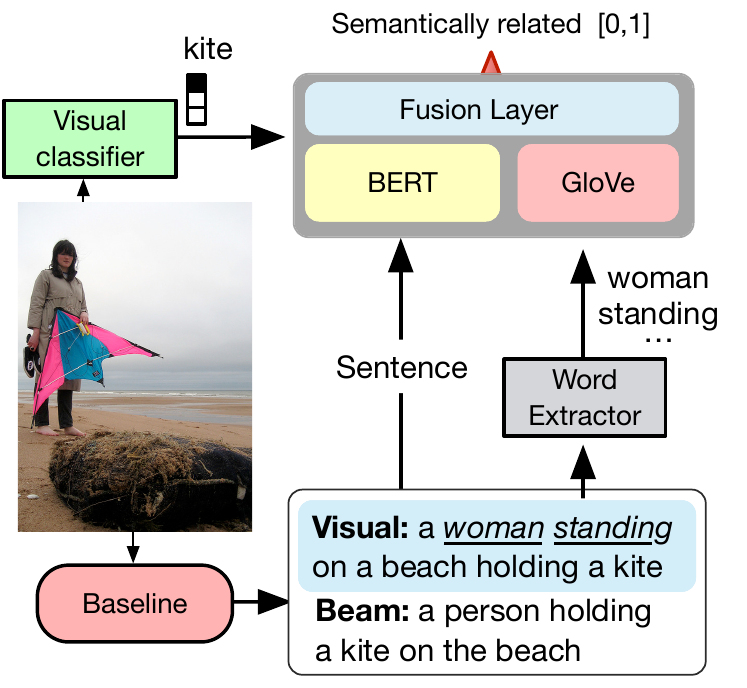}



\caption{An overview of our visual semantic re-ranker. We employ the visual context in a word and sentence level manner from the image to re-rank the most closely related caption to its visual context. An example from the caption Transformer \cite{Marcella:20} shows how the \textbf{Visual} re-ranker  uses the semantic relation to re-rank a more diverse caption.}

 
\label{fig:figure_1}
\end{figure}

Automatic caption generation is a fundamental task that incorporates vision and language. The task can be tackled in two stages: first, image-visual information extraction and then  linguistic description generation. Most models couple the relations between visual and linguistic information via a Convolutional Neural Network (CNN) to encode the input image and Long Short Term Memory for language  generation (LSTM) \cite{Oriol:15,anderson2018bottom,lu2018neural}. Recently, self-attention has been used to  learn these relations via Transformers \cite{huang2019attention, Marcella:20, pan2020x} or Transformer encoder based models like Vision and Language BERT \cite{lu2019vilbert, lu202012,li2020oscar}. These systems show promising results on benchmark datasets such as Flickr \cite{hodosh2013framing} and COCO \cite{Tsung-Yi:14}.  However, the lexical diversity\footnote{Lexical diversity is a measure that counts the different words (unique words) used in a sentence.} of the generated caption remains a relatively unexplored research problem. Lexical diversity refers to how accurate the generated description is for a given image. An accurate caption should provide details regarding specific and relevant aspects of the image \cite{luo2018discriminability}. Caption lexical diversity can be divided into three levels: word level (different words), syntactic level (word order), and semantic level (relevant concepts) \cite{wang2019describing}. In this work, we approach word level diversity at the semantic level by learning the semantic correlation between the caption and its visual context, as shown in Figure \ref{fig:figure_1}, where the visual information from the image is used  to learn the semantic relation from  the caption in  a word and sentence manner.




Modern image captioning systems heavily rely on visual grounding, using objects or regions from the image to guide the caption generation \cite{fang2015captions,  cornia2019show, zhou2020more}.  The informativeness of visual information will help the model to narrow the search (\ie beam search) to determine the most related candidate caption to the object or scene in the image. Inspired by these works,  we propose an object-based re-ranker to re-rank the most closely related caption with both static and contextualized semantic similarity.



 
\begin{table*}[t!]
\small 
\centering
\resizebox{0.7\linewidth}{!}{
\begin{NiceTabular}{llcccccccc}
\toprule   
 Model & B-1  & B-2 & B-3 & B-4 & M & R & C & BERTscore\\ 
\midrule

\multicolumn{10}{l}{Show and Tell CNN-LSTM \cite{Oriol:15} \text{\small$\spadesuit$}}\\ 


\addlinespace[0.1cm]


Tell$_{\text{BeamS}}$ & \textbf{0.331}  &   \textbf{0.159} & 0.071& 0.035 & 0.093 & 0.270 & 0.035 &  \textbf{0.8871}  \\ 

\addlinespace[0.1cm]
Tell+VR_V1$_{\text{BERT-GloVe}}$  & 0.330 & 0.158         &   0.069          &  0.035 &  0.095 & 0.273 & 0.036  &     0.8855 \\

Tell+VR_V2$_{\text{BERT-GloVe}}$& 0.320 & 0.154 & \textbf{0.073} &  \textbf{0.037} &0.099 & \textbf{0.277} & \textbf{0.041} & 0.8850  \\

Tell+VR_V1$_{\text{RoBERTa-GloVe}}$ (sts)   & 0.313 & 0.153 & 0.072 & \textbf{0.037} & \textbf{0.101} & 0.273 & 0.036 & 0.8839  \\
Tell+VR_V2$_{\text{RoBERTa-GloVe}}$ (sts) & 0.330 & 0.158 & 0.069 & 0.035 & 0.095 & 0.273 & 0.036  &  0.8869 \\

\midrule 
\addlinespace[0.1cm]

\multicolumn{8}{l}{VilBERT \cite{lu202012} \text{\small$\clubsuit$}}    \\

\addlinespace[0.1cm]
Vil$_{\text{BeamS}}$  &  0.739 &  0.577  & 0.440    & 0.336  & 0.271 & 0.543   & 1.027 & 0.9363  \\ 

\addlinespace[0.1cm]

Vil+VR_V1$_{\text{BERT-GloVe}}$   & 0.739  & 0.576   &  0.438 &  0.334    &  \textbf{0.273}   &0.544 &   1.034 &  0.9365 \\
Vil+VR_V2$_{\text{BERT-GloVe}}$  & \textbf{0.740} &  0.578  & 0.439   &  0.334 &  \textbf{0.273}    &   \textbf{0.545}  & 1.034 &   0.9365 &     \\

Vil+VR_V1$_{\text{RoBERTa-GloVe}}$ (sts) &  0.738 & 0.576  & 0.440   &  0.335 &  \textbf{0.273}    &  0.544    &   1.036    &   0.9365          \\
Vil+VR_V2$_{\text{RoBERTa-GloVe}}$ (sts)  & \textbf{0.740} & \textbf{0.579}  & \textbf{0.442}   &   \textbf{0.338} &  0.272    &  \textbf{0.545}    &  \textbf{1.040}   &   \textbf{0.9366}       \\


\midrule 
\addlinespace[0.1cm]
\multicolumn{8}{l}{Transformer based Caption Generator \cite{Marcella:20} \text{\small$\clubsuit$}}  \\

\addlinespace[0.1cm]
    Trans$_{\text{\text{BeamS}}}$    & \textbf{0.780} &  \textbf{0.631}  & \textbf{0.491}     &  \textbf{0.374}  & \textbf{0.278}   & \textbf{0.569}   & \textbf{1.153} & \textbf{0.9399}\\ 

\addlinespace[0.1cm]
Trans+VR_V1$_{\text{BERT-GloVe}}$   & \textbf{0.780} & 0.629   &  0.487 &  0.371    &  \textbf{0.278}   & 0.567 &   1.149 & 0.9398 \\

Trans+VR_V2$_{\text{BERT-GloVe}}$   & \textbf{0.780}  & 0.630   &  0.488 &  0.371   &  \textbf{0.278}     & 0.568 &   1.150  & \textbf{0.9399}  \\ 


Trans+VR_V1$_{\text{RoBERTa-GloVe}}$ (sts) & 0.779  & 0.629 & 0.487 & 0.370 & 0.277 & 0.567  & 1.145  & 0.9395   \\ 
Trans+VR_V2$_{\text{RoBERTa-GloVe}}$ (sts)   & 0.779 & 0.629 & 0.487 & 0.370 & 0.277  &  0.567 & 1.145 & 0.9395  \\ 



\bottomrule

\end{NiceTabular}
}
\vspace{0.5cm}
\caption{Performance of compared baselines on the 5k Karpathy test split \text{\small$\clubsuit$} (for VilBERT and Transformer baselines) and Flickr8k 1730 test set \text{\small$\spadesuit$} (for Show and Tell CNN-LSTM baseline) with/without \textbf{V}isual semantic \textbf{R}e-ranking. At inference, we use only Top-$k$-2 (\textbf{V}isual 1 or \textbf{V}isual 2) object visual context once at a time.  }

\label{se:Table}
\end{table*}

Our main contributions in this paper are: (1) we propose a post-processing method for any caption generation system via visual semantic related measures; (2) as an addendum to the main analysis of this work, we note that the visual re-ranker does not apply to less diverse short beam search, and suffer from the fluctuating of independent stand-alone word similarity score.

\section{Beam Search Caption Extraction}
\vspace{-0.3cm}


We employ the three most common architectures for caption generation to extract the top beam search. The first baseline is based on the standard CNN-LSTM model \cite{Oriol:15}. The second,  VilBERT \cite{lu202012}, is fine-tuned on a total of 12 different vision and language datasets such as caption image retrieval. Finally, the third baseline is a specialized Transformer caption generator \cite{Marcella:20}.



\section{Visual Re-ranking for Image Captioning}
\vspace{-0.2cm}

\noindent{\textbf{Problem Formulation.}} Beam search is the dominant method for approximate decoding in structured prediction tasks such as machine translation and image captioning. The larger beam size allows the model to perform a better exploration of the search space compared to greedy decoding.  Our goal is to leverage the visual context information of the image to re-rank the candidate sequences obtained through the beam search, thereby moving the most visually relevant candidate up in the list, while moving incorrect candidates down.

\begin{figure*}[t!]
\centering 

\includegraphics[width=\textwidth]{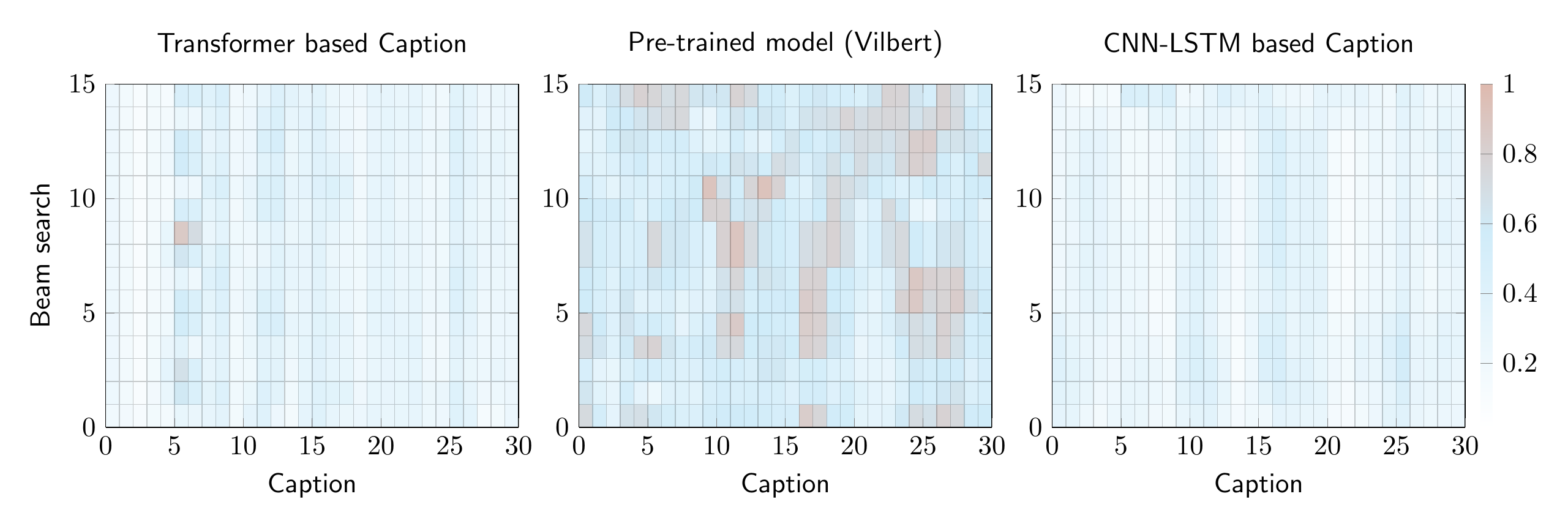}

\vspace{-0.3cm}

\caption{Visualization of the top-15 beam search after visual re-ranking. The color $\square$ $\leq 0$, \textcolor{babyblue!40}{$\blacksquare$} $\leq 0.4$ and  $\leq 0.8$ \textcolor{darksalmon}{$\blacksquare$} represents  the degree of change in probability after visual re-ranking, respectively.  Also, we can observe that a less diverse beam negatively impacted the score, as in the case of Transformer and CNN-LSTM baselines. }


\label{fig:muti}
\end{figure*} 

\noindent{\textbf{Beam Search Visual Re-ranking.}} We introduce a word and sentence level semantic relation with the visual context in the image. Inspired by \cite{peinelt2020tbert}, who propose a joint BERT \cite{Jacob:19} with topic modelling for semantic similarity, we propose a joint BERT with GloVe for visual semantic similarity.






\noindent{\bf Word Level Similarity.} To learn the semantic relation between a caption and its visual context in a word level manner, we first employ a bidirectional LSTM based CopyRNN keyphrase extractor \cite{meng2017deep} to extract keyphrases from the sentence as context. The model is trained on two combined pre-processed datasets: (1) wikidump (\ie  keyword, short sentence) and (2) SemEval 2017 Task 10 (Keyphrases from scientific publications) \cite{augenstein2017semeval}. Secondly, GloVe \cite{pennington2014glove} is used to compute the cosine similarity between the  visual context and its related context. Lastly, following \cite{sabir2019visual} we use the object confidence score (\ie visual context) in the image to convert the similarity to probability \cite{blok2003probability}. For example, from \textit{a woman in a red dress and a black skirt walks down a sidewalk} the model will extract \textit{dress}  and \textit{walks}, which are the highlights keywords of the caption.

\noindent{\bf Sentence Level Similarity.} We fine-tune the BERT base model to learn the visual context information. The model learns a dictionary-like relation word-to-sentence paradigm. We use the visual data (\ie object as context for the caption) to compute the relatedness score.


\begin{itemize}
   \itemsep 0cm


\item \textbf{BERT.} BERT achieves remarkable results on many sentence level tasks and especially in the  Semantic Textual Similarity task (STS-B) \cite{cer2017semeval}. Therefore, we fine-tuned BERT$_{\text{base}}$ on the training dataset, (textual information, 460k captions: 373k for training and 87k for validation) \ie visual, caption, label ([semantically related/not related]), with a binary classification cross-entropy loss function [0,1] where the target is the semantic similarity between the visual and the candidate caption, with a batch size of 16 for 2 epochs.

\item \textbf{RoBERTa \cite{liu2019roberta}.}  RoBERTa is an improved version of BERT, and since RoBERTa$_{\text{Large}}$ is more robust, we rely on pre-trained \textbf{S}entence RoBERTa-sts  \cite{reimers2019sentence} that are fine-tuned on general STS-B dataset \cite{cer2017semeval}.

\end{itemize}

\noindent{\bf Fusion Similarity Expert.} Inspired by Product of Experts PoE \cite{hinton1999products}, we combined the two experts at the word and sentence level  as a Late Fusion layer as shown in Figure \ref{fig:figure_1}. The combined probability of a given candidate caption $\mathbf{w}$, with the semantic relation with the visual context, can be
written as:




 \begin{small}
\begin{equation}
P\left(\mathbf{w} | \theta_{1} \ldots \theta_{n}\right)=\frac{\Pi_{m} p_{m}\left(\mathbf{w} | \theta_{m}\right)}{\sum_{\mathbf{c}} \Pi_{m} p_{m}\left(\mathbf{c} | \theta_{m}\right)}
\end{equation}
\end{small}




\noindent{where} $\theta_{m}$ are the parameters of each model $m$, $p_{m}\left(\mathbf{w} \mid \theta_{m}\right)$ is the probability of $\mathbf{w}$ under the model $m$, and $c$ is the indexes of all possible vectors in the data space. Since we are just interested in retrieving the candidate caption with higher probability after re-ranking, we do not need to normalize. Therefore, we compute:

\begin{small}
\begin{align}
\arg \max _{\mathbf{w}} P\left(\mathbf{w} | \theta_{1} \ldots \theta_{n}\right)&= \arg  \max _{\mathbf{w}} {\Pi_{m} p_{m}\left(\mathbf{w} | \theta_{m}\right)} 
\end{align}
\end{small}

\noindent{where},  in our case, $p_{m}\left(\mathbf{w} | \theta_{m}\right)$ are the probabilities (\ie semantic relatedness score) assigned by each expert to the candidate caption $\mathbf{w}$ with the semantic relation with the visual context of the image.





\section{Datasets}
\vspace{-0.4cm}

We evaluate the proposed approach on two different size datasets. The idea is to evaluate our approach with  (1) a shallow model CNN-LSTM  (\ie less data scenario), and on  a system that is trained on a huge amount of data (\ie Transformer).

\noindent{\bf Flickr8k \cite{hodosh2013framing}.} The dataset contains 8k images, and each image has five human label annotated captions. We use this data to train the shallow model CNN-LSTM baseline (6270 train/1730 test).


\noindent{\bf  COCO \cite{Tsung-Yi:14}.}  It contains around 120k images, and each image is annotated with five different human label captions. We use the most commonly used split as provided by  Karpathy \textit{et al.} \cite{karpathy2015deep}, where 5k images are used for testing and 5k for validation, and the rest for model training for the Transformer baseline.



\noindent{\bf{Visual Context Dataset.}} Since there are many public datasets for image captioning, they contain no visual information like objects in the image. We enrich the two datasets, as mentioned above, with textual visual context information. In particular, to automate visual context generation and dispense with the need for human labeling, we use ResNet-152 \cite{Kaiming:16} to extract top-$k$ 3 visual context from each image in the caption dataset. 



%


\begin{table}[t!]\centering

\small 
\resizebox{0.8\columnwidth}{!}{
\begin{tabular}{lcccc}

\hline 
Model & Voc & TTR &  Uniq & WPC\\
\hline 
\addlinespace[0.1cm]

\multicolumn{4}{l}{Show and Tell CNN-LSTM \cite{Oriol:15} \text{\small$\spadesuit$}}  \\ 

\addlinespace[0.1cm]

Tell$_{\text{BeamS}}$  &     304     	& 0.79	 &  \textbf{10.4}	&  12.7  \\
\addlinespace[0.1cm]
Tell+VR$_{\text{RoBERTa-GloVe}}$     & \textbf{310} & \textbf{0.82}	 & 9.42& \textbf{13.5}  \\	 \addlinespace[0.1cm]
\hline \hline
 \addlinespace[0.1cm]
\multicolumn{4}{l}{VilBERT \cite{lu202012} \text{\small$\clubsuit$} }	 \\ 
\addlinespace[0.1cm]
Vil$_{\text{BeamS}}$  &     894     	&  \textbf{0.87}	 &  8.05	&  10.5    \\
\addlinespace[0.1cm]
Vil+VR$_{\text{RoBERTa-GloVe}}$     & \textbf{953}  &  0.85	 & \textbf{8.86} & \textbf{10.8} \\	 \hline  
\addlinespace[0.1cm]

\multicolumn{4}{l}{Transformer based Caption Generator \cite{Marcella:20} \text{\small$\clubsuit$} } \\ 

\addlinespace[0.1cm]
Trans$_{\text{\text{BeamS}}}$   & 935 	& 	0.86  &7.44  & \textbf{9.62} \\ 
\addlinespace[0.1cm]
Trans+VR$_{\text{BERT-GloVe}}$   &	\textbf{936} & 	0.86   & \textbf{7.48} & 8.68 \\
\hline 

\end{tabular}
}
\vspace{0.5cm}
\caption{Measuring the lexical diversity of caption before and after re-ranking. Uniq and WPC columns indicate the average of unique/total Words Per Caption, respectively. (The  $\small \spadesuit$ refers to the Flickr8k 1730 test set, and  $\small \clubsuit$ refers to the  Karpathy 
5k test split on COCO Caption dataset.}  


\label{tb:lexical diversity}
\end{table}

{\bf{\noindent{Evaluation Metric}.}} We use the official COCO offline evaluation suite, producing several widely used caption quality metrics: \textbf{B}LEU \cite{papineni2002bleu} \textbf{M}ETEOR \cite{banerjee2005meteor}, \textbf{R}OUGE  \cite{lin2004rouge}, \textbf{C}IDEr \cite{vedantam2015cider} and BERTscore or (\textbf{B-S}) \cite{bert-score}.

\section{Experiments}

\vspace{-0.3cm}



We use visual semantic information to re-rank the candidate captions produced by out-of-the-box state-of-the-art caption generators. We extracted the top-20 beam search candidate captions from three different architectures (1) standard CNN-LSTM model \cite{Oriol:15}, (2)  a pre-trained  vision and language model VilBERT \cite{lu202012}, fine-tuned on a total of 12 different vision and language datasets such as caption image retrieval, and (3) a specialized caption-based Transformer \cite{Marcella:20}.


Experiments applying different re-rankers to each base system are shown in Table~\ref{se:Table}. 
The tested re-rankers are: (1) VR$_{\text{BERT-GloVe}}$, which uses BERT and GloVe similarity between the candidate caption and the visual context (top-$k$ V1 or V2 during the inference) to obtain the re-ranked score. (2) VR$_{\text{RoBERTa-GloVe}}$, which carries out the same procedure using SRoBERTa.

Our re-ranker produced mixed results as the model struggles when the beam search is less diverse. The model is therefore not able to select the most closely related caption to its environmental context as shown in Figure \ref{fig:muti}, which is a visualization of the final visual beam re-ranking. However, our models improve the lexical diversity, as shown in Table \ref{tb:lexical diversity} and we can conclude that we have (1) more Vocabulary and (2) the Unique words/total Words Per Caption are also improved, even with a lower Type-Token Ratio (TTR \cite{brown2005encyclopedia}\footnote{TTR is the number of unique words or types divided by the total number of tokens in a text fragment.}).

\begin{figure}[t!]
\small

\resizebox{\columnwidth}{!}{
 \begin{tabular}{C{3.3cm}  L{5cm}}

      \includegraphics[width=\linewidth, height = 2.5 cm]{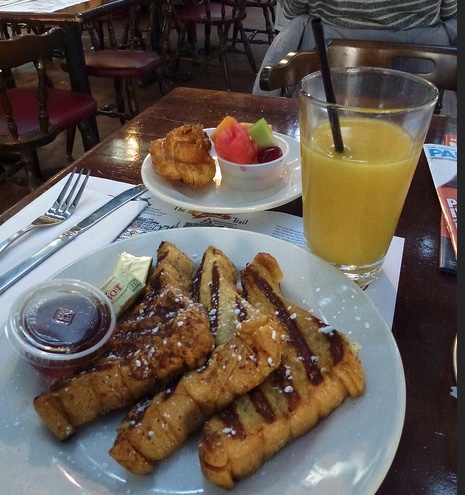} Visual:  food &  \textbf{BL$_{\text{BeamS}}$:}  a plate of food on a table \newline 
         \textbf{VR$_{\text{BERT-GloVe}}$:}  a plate of food \textcolor{blue}{and a drink} on a table
    \newline \textbf{Human:} a white plate with some \newline food on it. \\
         
  
     \arrayrulecolor{gray}\hline 
       \vspace{-0.2cm}


      \includegraphics[width=\linewidth, height = 2.5 cm]{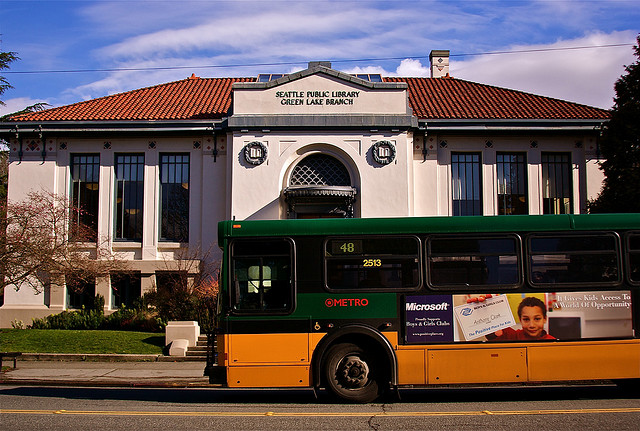}  Visual: trolleybus 
  &  \textbf{BL$_{\text{greedy}}$:}   a green bus parked in front  of a building \newline
         \textbf{VR$_{\text{BERT-GloVe}}$:}   a green double decker  bus   parked in  front  of a building {\color{red}{\xmark}} \newline
 \textbf{Human:} a passenger bus that is  parked in front of a library. \\  

        \\ 

    \end{tabular}
}
 \vspace{-0.2cm}

\caption{Examples of the re-ranked captions by our visual re-ranker (VR) against the original caption (\textbf{greedy} and \textbf{Beam Search}) by the baseline (BL). The (Top) example shows that our model re-ranked a more diverse caption than the baseline. (Bottom) the model struggles when there is a high similarity word with the visual context and without direct relation to the image \ie $sim$(trolleybus, decker) which influences the final score negatively.}

\label{tb-fig:example_1}

\end{figure}

Figure \ref{tb-fig:example_1} shows examples of the re-ranked captions by our visual re-ranker (VR) against greedy/beam search. (Top) the re-ranked caption has a precise description of the image \textit{food and drink}. In the (Bottom) example, the independent word level (\ie single word without surrounding context) high similarity score   \textit{sim}(trolleybus, decker) influences the expert decision negatively,  which results in an incorrect re-ranked caption as the object is not present in the image.




\begin{figure}[t!]
\centering 

\includegraphics[width=0.8\columnwidth]{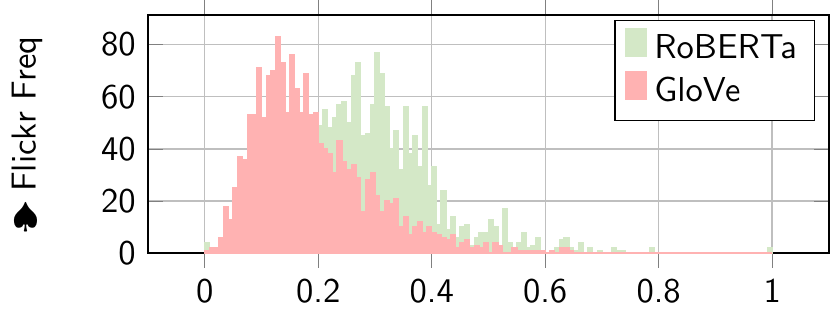}
\includegraphics[width=0.8\columnwidth]{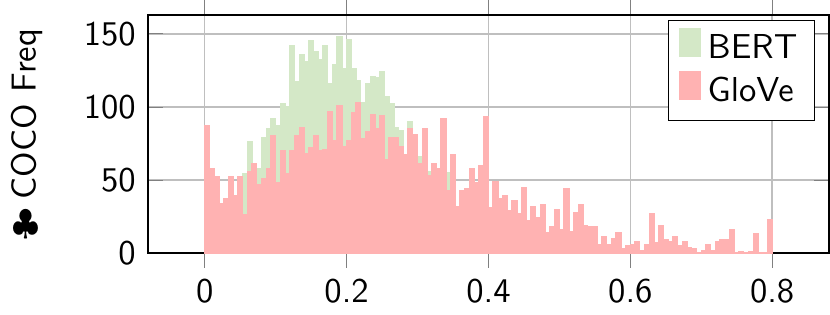}

\vspace{0.1cm}


\caption{(\textbf{\text{\small$\spadesuit$} Top}) 1k random sample from Flickr test set with CNN-LSTM baseline. Each Expert is contributing different probability confidence and therefore the model is learning the semantic relation in word and sentence level. (\text{\small$\clubsuit$} \textbf{Bottom}) 5k Karpathy test split from COCO Caption with caption Transformer baseline. The GloVe score is dominating the distribution to become the expert.  }

 
\label{fig:Glove-BERT}
\end{figure} 

\begin{table}[t!]\centering

\small 
\resizebox{\columnwidth}{!}{
\begin{tabular}{lcccccc}

\toprule
 Model &  B-4 & M & R & C  & B-S \\
\midrule 
\multicolumn{5}{l}{Transformer based Caption Generator \cite{Marcella:20}}  \\
\addlinespace[0.1cm]
Trans$_{\text{\text{BeamS}}}$  &  \textbf{0.374}& \textbf{0.278} & \textbf{0.569} & \textbf{1.153} & \textbf{0.9399} \\ 
\addlinespace[0.1cm]

+VR$_{\text{\text{RoBERT-GloVe}}}$ & 0.370 & 0.277 & 0.567  & 1.145  & 0.9395   \\ 
+VR$_{\text{\text{BERT-GloVe}}}$     & \cellcolor{red!30}0.371    & \cellcolor{red!30}0.278   & 0.567 & \cellcolor{red!30}   1.149 & 0.9398  \\ 
 
+VR$_{\text{\text{RoBERT-BERT}}}$ &  0.369 & 0.278  & 0.567 & 1.144 & 0.9395 \\
\hline
+VR_V1$_{\text{\text{GloVe}}}$   &  0.371   & 0.278 &   0.568 &  1.148 & 0.9398  \\
+VR_V2$_{\text{\text{GloVe}}}$  &  \cellcolor{green!40}0.371  & \cellcolor{green!40}0.278  &  0.568 & \cellcolor{green!40}1.149 & 0.9398 \\ 
\bottomrule

\end{tabular}
}

\vspace{0.6cm}

\caption{Ablation study using different model compared to GloVe alone  visual re-ranker on the Transformer baseline. \text{\small$\clubsuit$} (Bottom) Figure \ref{fig:Glove-BERT}  shows that \textcolor{red!30}{BERT} is not contributing, as \textcolor{green!70}{GloVe}, to the final score for two reasons: (1) less diverse short beam, and (2) the fluctuating of independent word level  similarity score as shown in Figure \ref{tb-fig:example_1} (Bottom).}


\label{tb:ablation}
\end{table}


\noindent{\textbf{Ablation Study.}} We performed an ablation study, with our worst model in Table \ref{se:Table} \ie caption Transformer, to investigate the effectiveness of each model (\ie word and sentence experts). In this experiment, we trained  each  model separately, as shown in Table \ref{tb:ablation}. The GloVe performed better as a stand-alone than the combined model (and thus, the combined model breaks the accuracy). To investigate this even further we visualized each expert before the fusion layers as shown in Figure \ref{fig:Glove-BERT} (Bottom), BERT struggles with the Transformer less  diverse short caption, and  therefore, the word level \ie GloVe dominates as  the main expert.




\noindent{\textbf{Limitation.}} In contrast to CNN-LSTM  Figure \ref{fig:Glove-BERT} (Top), where each expert is contributing to the final decisions, we observed that having a shorter caption (with less context) can influence the BERT similarity score negatively.  Another limitation is the fluctuating of independent stand-alone word similarity score \ie keyphrases from the caption and the visual $sim$(keyphrase, visual) as shown in the Figure \ref{tb-fig:example_1} (Bottom) with \textit{sim}(trolleybus, decker) example. Also, the visual classifier struggles with complex backgrounds (\ie misclassified/hallucinated objects), which results in an inaccurate semantic score.

\section*{Conclusion}
\vspace{-0.3cm}


In this work, we have introduced an approach that overcomes the limitation of beam search and avoids re-training for better accuracy. We proposed a combined word and sentence visual beam search re-ranker. However, we discovered that word and sentence similarity disagree with each other when the beam search is less diverse. Our experiments also highlight the usefulness of the proposed model by showing successful cases.


\bibliographystyle{ieee_fullname}
\bibliography{custom}





\end{document}